\documentclass[runningheads]{llncs}
\usepackage[T1]{fontenc}
\usepackage{graphicx}
\usepackage{epstopdf}
\usepackage{amssymb}
\usepackage{amsmath}
\usepackage{marvosym}
\usepackage{multirow}
\usepackage{hyperref}
\usepackage{color}

\begin{document}
\title{Few-Shot Medical Image Segmentation via a Region-enhanced Prototypical Transformer}

\author{Yazhou Zhu\inst{1} \and  
Shidong Wang\inst{2} \and  
Tong Xin\inst{3}	\and 
Haofeng Zhang\inst{1} 
\textsuperscript{(\Letter)}} 
\authorrunning{Y. Zhu et al.}
%
\institute{School of Computer Science and Engineering, Nanjing University of Science and Technology, Nanjing 210094, China. 
\email{\{zyz\_nj,zhanghf\}@njust.edu.cn} \and
School of Engineering, Newcastle University, Newcastle upon Tyne, NE17RU, UK. \email{shidong.wang@newcastle.ac.uk} \and
School of Computing, Newcastle University, Newcastle upon Tyne, NE17RU, UK. \email{tong.xin@newcastle.ac.uk}}

\maketitle              
\begin{abstract}

Automated segmentation of large volumes of medical images is often plagued by the limited availability of fully annotated data and the diversity of organ surface properties resulting from the use of different acquisition protocols for different patients. In this paper, we introduce a more promising few-shot learning-based method named Region-enhanced Prototypical Transformer (RPT) to mitigate the effects of large intra-class diversity/bias. First, a subdivision strategy is introduced to produce a collection of regional prototypes from the foreground of the support prototype. Second, a self-selection mechanism is proposed to incorporate into the Bias-alleviated Transformer (BaT) block to suppress or remove interferences present in the query prototype and regional support prototypes. By stacking BaT blocks, the proposed RPT can iteratively optimize the generated regional prototypes and finally produce rectified and more accurate global prototypes for Few-Shot Medical Image Segmentation (FSMS). Extensive experiments are conducted on three publicly available medical image datasets, and the obtained results show consistent improvements compared to state-of-the-art FSMS methods. The source code is available at: \href{https://github.com/YazhouZhu19/RPT}{https://github.com/YazhouZhu19/RPT}.
\end{abstract}
\section{Introduction}
Automatic medical image segmentation is the implementation of data-driven image segmentation concepts to identify a specific anatomical structure's surface or volume in a medical image ranging from X-ray and ultrasonography to CT and MRI scans. Deep learning algorithms are exquisitely suited for this task because they can generate measurements and segmentations from medical images without the time-consuming manual work as in traditional methods. However, the performance of deep learning algorithms depends heavily on the availability of large-scale, high-quality, fully pixel-wise annotations, which are often expensive to acquire. To this end, few-shot learning is considered as a more promising approach and introduced into the medical image segmentation by \cite{ouyang2020self}.

\begin{figure*}[!t]
\centering
\includegraphics[width=0.9\textwidth]{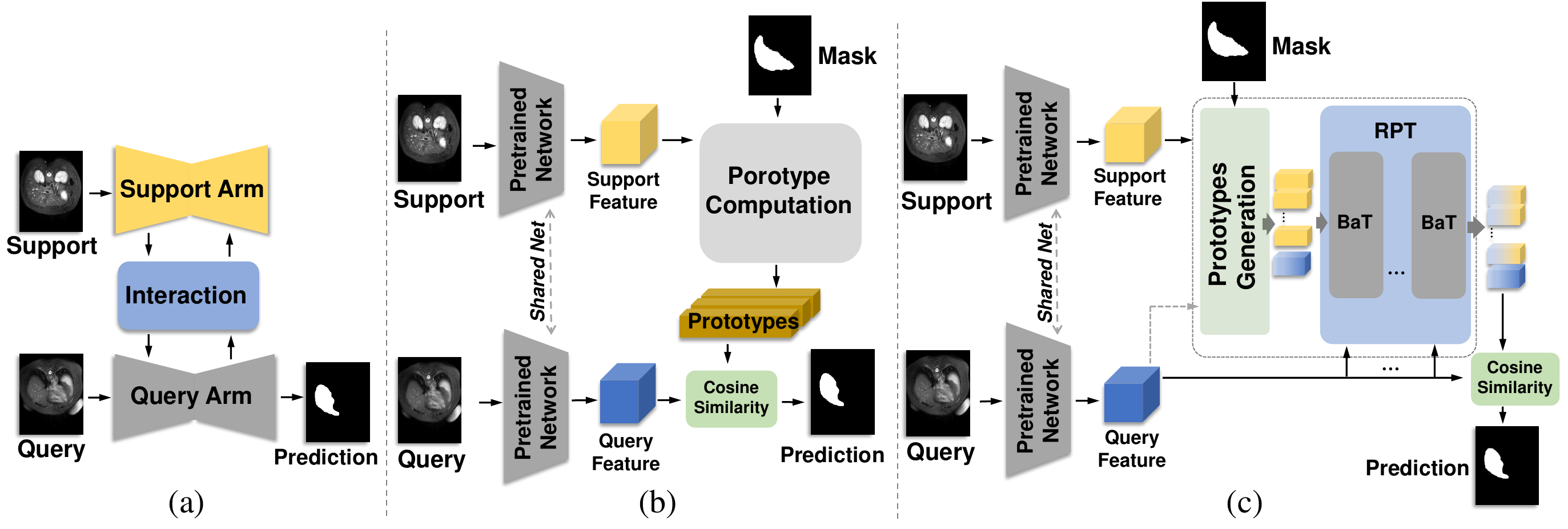}
\vspace{-3ex}
\caption{Comparison between previous FSMS models and our model. (a) Interactive model. (b) Prototypical network based model. (c) Our proposed model.}
\label{fig1}
\vspace{-4ex}
\end{figure*}

Through revisiting existing FSMS algorithms \cite{ding2023few,feng2021interactive,hansen2022anomaly,shen2021poissonseg,shen2022q,sun2022few}, they can be grouped into two folders, including the interactive method originated from SENet \cite{roy2020squeeze} (shown in Fig. \ref{fig1}(a)) and the prototype networks \cite{snell2017prototypical,wang2019panet} (demonstrated in Fig. \ref{fig1}(b)). For the interaction-based approach, the ideas of \textit{attention} \cite{sun2022few}, and \textit{contrastive learning} \cite{wu2022dual} are introduced to work interactively between parallel support and query arms. In contrast, prototype network-based approach almost dominates the FSMS research, such as SSL-ALPNet \cite{ouyang2020self}, ADNet \cite{hansen2022anomaly} and SR\&CL~\cite{wang2022few}, whose core idea is to obtain semantic-level prototypes by compressing support features, and then make predictions by matching with query features. However, the problem of how to obtain an accurate and representative prototype remains. 

The main reason affecting the representativeness of the prototype is the significant discrepancy between support and query. Specifically, in general, different protocols are taken for different patients, which results in a variety of superficial organ appearances, including the \textit{size}, \textit{shape}, and \textit{contour} of features. In this case, the prototype generated from the support features may not accurately represent the key attributes of the target organ in the query image. In addition, it is also challenging to extract useful information (prototypes of novel classes) from the cluttered background due to the extremely heterogeneous texture between the target and its surroundings, which may contain information belonging to some novel classes or redundant information issue \cite{sun2022few}.

To mitigate the impact of intra-class diversity, it considers subdividing the foreground of the supporting prototypes to produce some regional prototypes, which are then rectified to suppress or exclude areas inconsistent with the query targets, as illustrated in Fig. \ref{fig1}(c). Concretely, in the prototype learning stage, multiple subdivided regional prototypes are enhanced with a more accurate class center, which can be derived from the newly designed Regional Prototype Generation (RPG) and Query Prototype Generation (QPG) modules. Then, a designed Region-enhanced Prototypical Transformer (RPT) that is mainly composed of a number of stacked Bias-alleviated Transformer (BaT) blocks, each of which contains the core debiasing function-Search and Filter (S\&F) modules, to filter out undesirable prototypes. As shown in Fig. \ref{fig2},  Our contributions are summarized as follows:     
\begin{itemize}
    \item A Region-enhanced Prototypical Transformer (RPT) consisting of stacked Bias-alleviated Transformer (BaT) blocks is proposed to mitigate the effects of large intra-class variations present in FSMS through Search and Filter (S\&F) modules devised based on the self-selection mechanism.
    \item A subdivision strategy is proposed to perform in the foreground of the support prototype to generate multiple regional prototypes, which can be further iteratively optimized by the RPT to produce the optimal prototype.
    \item The proposed method can achieve state-of-the-art performance on three experimental datasets commonly used in medical image segmentation.
\end{itemize}

\section{Methodology}

\subsection{Overall Architecture}
Before introducing the overall architecture, it is necessary to briefly explain how data is processed. Specifically, the 3D supervoxel clustering method \cite{hansen2022anomaly} is employed to generate pseudo-masks as supervision, which is learned in a self-supervised learning manner without any manual annotations. Meta-learning-based episodic tasks can then be constructed using the generated pseudo-masks. Notably, the pseudo-masks obtained by the 3D clustering method is more consistent with the volumetric properties of medical images than the 2D superpixel clustering method adopted in \cite{ouyang2020self}.

\begin{figure*}[!t]
\centering
\includegraphics[width=0.98\textwidth, keepaspectratio]{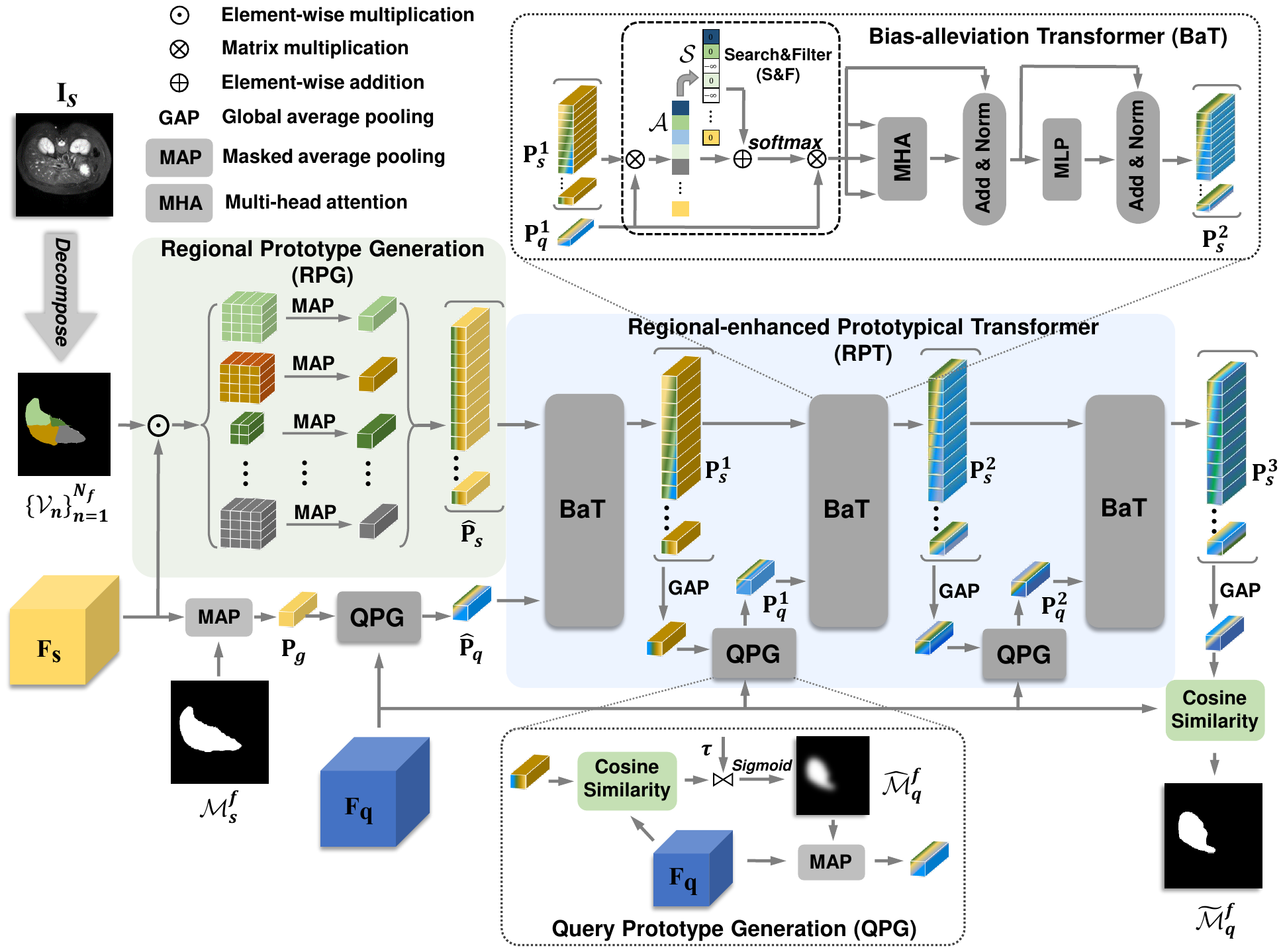}
\vspace{-3ex}
\caption{Overview of the proposed Region-enhanced Prototypical Transformer.}
\label{fig2}
\vspace{-3ex}
\end{figure*}

As depicted in Fig. \ref{fig2}, the overall architecture includes three main components: the Regional Prototype Generation (RPG) module, the Query Prototype Generation (QPG) module and the Region-enhanced Prototypical Transformer (RPT) consisting of three Bias-alleviated Transformer (BaT) blocks. The pipeline first extracts features from support and query images using a weight-shared ResNet-101 \cite{he2016deep} as a backbone, which has been pretrained on the MS-COCO dataset \cite{lin2014microsoft}. We employ the ResNet101 pretrained on MS-COCO for optimal performance, and the comparison with ResNet50 pretrained on ImageNet dataset\cite{deng2009imagenet} is also included in the appendix. The extracted features are then taken as the input of the RPG and QPG modules to generate multiple region prototypes, which will be rectified by the following RPT to produce the optimal prototype.

\subsection{Regional Prototype Generation}

The core problem considered in this paper is what causes prototype bias. By examining the input data, it can be observed that images of healthy and diseased organs have a chance to be considered as support or query. This means that if there are lesioned or edematous regions in some areas of the support images, they will be regarded as biased information which in reality cannot be accurately transferred for the query images containing only healthy organs. When these prototypes that contain the natural heterogeneity of the input images are processed by the Masked Average Pooling (MAP) operation, they inevitably lead to significant intra-class biases. 

To cope with the above problems, we propose a Region Prototype Generation (RPG) module to generate multi-region prototypes by performing subdivisions in the foreground of the support images. Given an input support image $\textbf{I}_{s}$ and the corresponding foreground mask $\mathcal{M}^{f}$, the foreground of this image can be obtained by calculating their product. The foreground image then can be partitioned into $N_f$ regions, where $N_f$ is set to 10 by default. By using the Voronoi-based partition method \cite{aurenhammer1991voronoi,zhang2022feature}, a set of regional masks $\left \{ \mathcal{V}_n \right \}^{N_f}_{n=1} $ can be derived for subsequent use of Masked Average Pooling (MAP) to generate a set of coarse regional prototypes $\hat{\mathcal{P}}_{s} = \left \{  \hat{p}_{n}\right \}^{N_f}_{n=1}, \hat{p}_{n} \in \mathbb{R}^{C}$. Formally, 
\begin{equation}\label{eq1}
\hat{p}_{n} = \operatorname{MAP}(\textbf{F}_{s}, \mathcal{V}_{n}) = \frac{1}{\left |  \mathcal{V}_n\right | } \sum_{i=1}^{HW}\textbf{F}_{s,i}\mathcal{V}_{n,i},
\end{equation}
where $\textbf{F}_{s}\in \mathbb{R}^{C \times H \times W}$ is the feature extracted from the support images and $\mathcal{V}_{n}$ denotes the regional masks. 
\subsection{Query Prototype Generation}

Once a set of coarse regional prototypes $\hat{\mathcal{P}}_s$ have been generated for the support images, we can employ the method introduced in \cite{liu2020prototype} to learn the coarse query prototype $\hat{\mathbf{P}}_{q} \in \mathbb{R}^{1 \times C}$. Concretely, it first uses the $\operatorname{MAP}(\cdot)$ operator as introduced in Eq. (\ref{eq1}) to learn a global support prototype $\mathbf{P}_g = \operatorname{MAP}(\textbf{F}_{s}, \mathcal{M}_{s})$ with $\mathbf{P}_{g} \in \mathbb{R}^{1 \times C}$, whose output can then be used to calculate the coarse query foreground mask $\hat{\mathcal{M}}^{f}_{q}$. Considering that the empirically designed threshold described in \cite{liu2020prototype} may affect the quality of the $\hat{\mathcal{M}}^{f}_{q}$, we hereby introduce a learnable threshold $\tau $. This process can be denoted as     
\begin{equation}\label{eq2}
    \hat{\mathcal{M}}_{q}^{f} = 1 - \sigma(S(\textbf{F}_{q}, \mathbf{P}_{g}) - \tau),
\end{equation}
where $\textbf{F}_q \in \mathbb{R}^{C \times H \times W}$ is feature extracted from query images, $S(a,b) = - \alpha cos(a,b)$ is the negative cosine similarity with a fixed scaling factor $\alpha = 20$, $\sigma $ denotes the Sigmoid activation, and $\tau$ is obtained by applying one average-pooling and two fully-connected layers ($\operatorname{FC} $) to the query feature, expressed as $\tau =  \operatorname{FC}(\textbf{F}_q)$. After this, the coarse query foreground prototype can be achieved by using $\hat{\mathbf{P}}_{q} = \operatorname{MAP}(\textbf{F}_{q, i}, \hat{\mathcal{M}}_{q, i}^{f})$.

\subsection{Region-enhanced Prototypical Transformer}
The above received prototypes $\hat{\mathcal{P}}_{s}$ and $\hat{\mathbf{P}}_{q}$ are taken as input to the proposed Region-enhanced Prototypical Transformer (RPT) to rectify and regenerate the optimal global prototype $\mathbf{P}_{s}$. As shown in Fig. \ref{fig2}, our RPT mainly consists of $L$ stacked Bias-alleviated Transformer (BaT) blocks each of which contains a Search and Filter (S\&F) module, and QPG modules that maintain the query prototypes continuously updated. Taking the first BaT block as an example, it calculates an affinity map $ \mathcal{A} = \hat{\mathbf{P}}_s \hat{\mathbf{P}}_{q}^{\top} \in \mathbb{R}^{N_f \times 1}$ to reveal the correspondence between the query and $N_f$ support regional prototypes by taking an input containing the query prototype $\hat{\mathbf{P}}_{q}$ and the support prototype $\hat{\mathbf{P}}_{s} \in \mathbb{R}^{N_f \times C}$ obtained by concatenating all elements in $\hat{\mathcal{P}}_{s}$. Then, a selective map $\mathcal{S} \in \mathbb{R}^{N_f \times 1}$ can be derived from the proposed self-selection based S\&F module by   
\begin{equation}
\mathcal{S}_{i}(\mathcal{A}_{i}) = \begin{cases}
             0 & \text{ if } \mathcal{A}_{i} >= \xi  \\
 -\infty  & otherwise 
\end{cases},   i \in  \left \{  0,1, ... , N_{f}\right \},
\end{equation}
where $\xi $ is the selection threshold achieved by $\xi = (min(\mathcal{A}) + mean(\mathcal{A}))/2$, $\mathcal{S}$ indicates the chosen regions from the support image that performs compatible with the query at the prototypical level. Then, the heterogeneous or disturbing regions of support foreground will be weeded out with $\operatorname{softmax}(\cdot)$ function. The preliminary rectified prototypes $\hat{\mathbf{P}}^{o}_{s} \in \mathbb{R}^{N_f \times C}$ is aggregated as:
\begin{equation}
\hat{\mathbf{P}}_{s}^{o} = \operatorname{softmax}(\hat{\mathbf{P}}_{s} \hat{\mathbf{P}}_{q}^{\top} + \mathcal{S})\hat{\mathbf{P}}_{q}.
\end{equation}

The refined $\hat{\mathbf{P}}_{s}^{o}$ will be fed into the following components designed based on the self-attention mechanism to produce the output $\mathbf{P}_{s}^{1}\in \mathbb{R}^{N_f \times C}$. Formally,
\begin{equation}
\hat{\mathbf{P}}^{o+1}_{s} = \operatorname{LN}(\operatorname{MHA}(\hat{\mathbf{P}}^{o}_{s}) + \hat{\mathbf{P}}^{o}_{s}), \qquad
\mathbf{P}_{s}^{1}=\operatorname{LN}(\operatorname{MLP}(\hat{\mathbf{P}}_{s}^{o+1})+\hat{\mathbf{P}}_{s}^{o+1}),
\end{equation}
where $\hat{\mathbf{P}}^{o+1}_{s} \in \mathbb{R}^{N_f \times C}$ is the intermediate generated prototype, $\operatorname{LN}(\cdot)$ denotes the layer normalization, $\operatorname{MHA}(\cdot)$ represents the standard multi-head attention module and $\operatorname{MLP}(\cdot)$ is the multilayer perception. 

By stacking multiple BaT blocks, our RPT can iteratively rectify and update all coarse support and the query prototype. Given the prototypes $\mathbf{P}^{l-1}_{s}$ and $\mathbf{P}^{l-1}_{q}$ from the previous BaT block, the updates for the current BaT block are computed by:
\begin{equation}
\mathbf{P}^{l}_{s}=\operatorname{BaT}(\mathbf{P}^{l-1}_{s}, \mathbf{P}^{l-1}_{q}), \qquad
\mathbf{P}^{l}_{q} = \operatorname{QPG}(\operatorname{GAP}(\mathbf{P}^{l}_{s}), \mathbf{F}_{q}),
\end{equation}
where $\mathbf{P}^{l}_s \in \mathbb{R}^{N_f \times C}$ and $\mathbf{P}^{l}_{q} \in \mathbb{R}^{1 \times C}$ ($l = 1,2,...,L$) are updated prototypes, $\operatorname{GAP}(\cdot)$ denotes the global average pooling operation. The final output prototypes $\mathbf{P}_{s}$ optimized by the RPT can be used to predict the foreground of the query image by using Eq.(\ref{eq2}: $\tilde{\mathcal{M}}^{f}_{q} = 1 - \sigma (S(\textbf{F}_{q}, \operatorname{GAP}(\textbf{P}^{3}_{s})) - \tau)$, while its background can be obtained by $\tilde{\mathcal{M}}^{b}_{q} = 1 - \tilde{\mathcal{M}}^{f}_{q}$ accordingly.  

\subsection{Objective Function}
The binary cross-entropy loss $\mathcal{L}_{ce}$ is adopted  to determine the error between the predict masks $\tilde{\mathcal{M}}_{q}$ and the given ground-truth $\mathcal{M}_{q} $. Formally,
\begin{equation}
\mathcal{L}_{ce} = -\frac{1}{HW}\sum_{h}^{H} \sum_{w}^{W} \mathcal{M}_q^{f}(x,y)log(\tilde{\mathcal{M}}_{q}^{f}(x, y)) + \mathcal{M}_q^{b}(x,y)log(\tilde{\mathcal{M}}_{q}^{b}(x, y)).
\end{equation}

Considering the prevalent class imbalance problem in medical image segmentation, the boundary loss \cite{kervadec2019boundary} $\mathcal{L}_{B}$ is also adopted and it is written as
\begin{equation}
\mathcal{L}_{B}(\theta ) = \int_{\Omega }\phi G(q) s_{\theta }(q)d_{q},
\end{equation} 
where $\theta $ denotes the network parameters, $\Omega$ denotes the spatial domain, $\phi G: \Omega \to \mathbb{R}$ denotes   the \textit{level set} representation of the ground-truth boundary, $\phi G(q) = -D_{G}(q) $ if $q \in G$ and $\phi G(q) = D_{G}(q)$ otherwise, $D_{G}$ is distance map between the boundary of prediction and ground-truth, and $s_{\theta }(q): \Omega  \to [0, 1]$ denotes $\operatorname{softmax}(\cdot)$ function. 

Overall, the loss used for training our RPT is defined as $\mathcal{L} = \mathcal{L}_{ce} + \eta \mathcal{L}_{dice} + (1 - \eta)\mathcal{L}_{B}$, where $\mathcal{L}_{dice}$ is the Dice loss \cite{ma2021loss}, $\eta$ is initially set to 1 and decreased by 0.01 every epoch.
\section{Experiments}
\noindent \textbf{Experimental Datasets:} The proposed method is comprehensively evaluated on three publicly available datasets, including \textbf{Abd-MRI}, \textbf{Abd-CT} and \textbf{Card-MRI}. Concretely, \textbf{Abd-MRI} \cite{kavur2021chaos} is an abdominal MRI dataset used in the ISBI 2019 Combined Healthy Abdominal Organ Segmentation Challenge. \textbf{Abd-CT} \cite{ABD-CT} is an abdominal CT dataset from MICCAI 2015 Multi-Atlas Abdomen Labeling Challenge. \textbf{Card-MRI}\cite{zhuang2018multivariate} is a cardiac MRI dataset from MICCAI 2019 Multi-Sequence Cardiac MRI Segmentation Challenge. All 3D scans are reformatted into 2D axial and 2D short-axis slices.  The abdominal datasets \textbf{Abd-MRI} and \textbf{Abd-CT} share the same categories of labels which includes the liver, spleen, left kidney (LK) and right kidney (RK). The labels for \textbf{Card-MRI} include left ventricular myocardium (LV-MYO), right ventricular myocardium (RV), and blood pool (LV-BP). 

\noindent \textbf{Experiment Setup:} The model is trained for 30k iterations with batch size set to 1. During training, the initial learning rate is set to $1 \times 10^{-3}$ with a step decay of $0.8$ every 1000 iterations. The values of $N_f$ and iterations $L$ are set to 10 and 3, respectively. To simulate the scarcity of labeled data in medical scenarios, all experiments embrace a 1-way 1-shot setting, and 5-fold cross-validation is also carried out in the experiments, where we only record the mean value.

\noindent \textbf{Evaluation:} For a fair comparison, the metric used to evaluate the performance of 2D slices on 3D volumetric ground-truth is the Dice score used in \cite{ouyang2020self}. Furthermore, two different supervision settings are used to evaluate the generalization ability of the proposed method: in Setting 1, the test classes may appear in the background of the training slices, while in Setting 2, the training slices containing the test classes are removed from the dataset to ensure that the test classes are unseen. Note that Setting 2 is impractical for Card-MRI scans, since all classes typically co-occur on one 2D slice, making label exclusion impossible. In addition, as in \cite{ouyang2020self}, abdominal organs are categorized into \textit{upper} abdomen (liver, spleen) and \textit{lower} abdomen (left, right kidney) to demonstrate whether the learned representations can encode spatial concepts.

\begin{table}[!t]
\centering
\caption{Quantitative Comparison (in Dice score $\%$) of different methods on abdominal datasets under \textit{Setting 1} and \textit{Setting 2}.}
\vspace{-2ex}
\label{tab1}
\resizebox{\textwidth}{!}{
\begin{tabular}{c|l|l|ccccc|ccccc}
\hline
\multicolumn{1}{c|}{} & \multicolumn{1}{c|}{} & \multicolumn{1}{c|}{} & \multicolumn{5}{c|}{Abd-MRI}        & \multicolumn{5}{c}{Abd-CT}           \\ 
\multicolumn{1}{c|}{Setting} & \multicolumn{1}{l|}{Method} & \multicolumn{1}{l|}{Reference} & \multicolumn{2}{c}{Lower} & \multicolumn{2}{c}{Upper} & \multirow{2}{*}{Mean} & \multicolumn{2}{c}{Lower} & \multicolumn{2}{c}{Upper} & \multirow{2}{*}{Mean} \\ 
\multicolumn{1}{c|}{} & \multicolumn{1}{c|}{} & \multicolumn{1}{c|}{}       & LK          & RK          & Spleen       & Liver      &                       & LK          & RK          & Spleen       & Liver      &                       \\ \hline

\multirow{5}{*}{1} &ADNet \cite{hansen2022anomaly} & MIA'22               & 73.86       & 85.80       & 72.29        & 82.11      & 78.51                 & 72.13       & 79.06      & 63.48        & 77.24      & 72.97                 \\
&AAS-DCL  \cite{wu2022dual}  & ECCV'22            & 80.37       & 86.11       & 76.24        & 72.33      & 78.76                 & 74.58       & 73.19       & 72.30        & 78.04      & 74.52                 \\
&SR\&CL \cite{wang2022few} & MICCAI'22           & 79.34       & 87.42       & 76.01        & 80.23      & 80.77                 & 73.45       & 71.22       & \textbf{73.41}        & 76.06      & 73.53                 \\
&CRAPNet \cite{ding2023few} & WACV'23            & \textbf{81.95}       & 86.42       & 74.32        & 76.46      & 79.79                 & 74.69       & 74.18       & 70.37        & 75.41      & 73.66                 \\
&\textbf{Ours (RPT)}  &      \multicolumn{1}{c|}{---}             & 80.72       & \textbf{89.82}       & \textbf{76.37}        & \textbf{82.86}      & \textbf{82.44}                 & \textbf{77.05}             & \textbf{79.13}            & 72.58            & \textbf{82.57}            & \textbf{77.83}                    \\ \hline
\multirow{5}{*}{2} & ADNet \cite{hansen2022anomaly} & MIA'22               & 59.64       & 56.68       & 59.44        & \textbf{77.03}      & 63.20                 & 48.41       & 40.52       & 50.97        & 70.63      & 52.63                 \\
&AAS-DCL  \cite{wu2022dual}  & ECCV'22            & 76.90       & 83.75       & 74.86        & 69.94     & 76.36                 & 64.71       & \textbf{69.95}       & 66.36        & 71.61      & 68.16                 \\
&SR\&CL \cite{wang2022few} & MICCAI'22           & 77.07       & 84.24       & 73.73        & 75.55      & 77.65                 & 67.39       & 63.37       & 67.36        & 73.63      & 67.94                 \\
&CRAPNet \cite{ding2023few} & WACV'23            & 74.66       & 82.77       & 70.82        & 73.82      & 75.52                 & 70.91       & 67.33       & 70.17        & 70.45      & 69.72                 \\
&\textbf{Ours (RPT)}  & \multicolumn{1}{c|}{---}                       & \textbf{78.33}       & \textbf{86.01}       & \textbf{75.46}        & 76.37      & \textbf{79.04}                 & \textbf{72.99}             & 67.73             & \textbf{70.80}             & \textbf{75.24}            & \textbf{71.69}                     \\ \hline
\end{tabular}}
\end{table}

\begin{figure}[!t]
\centering
\includegraphics[width=0.93\textwidth]{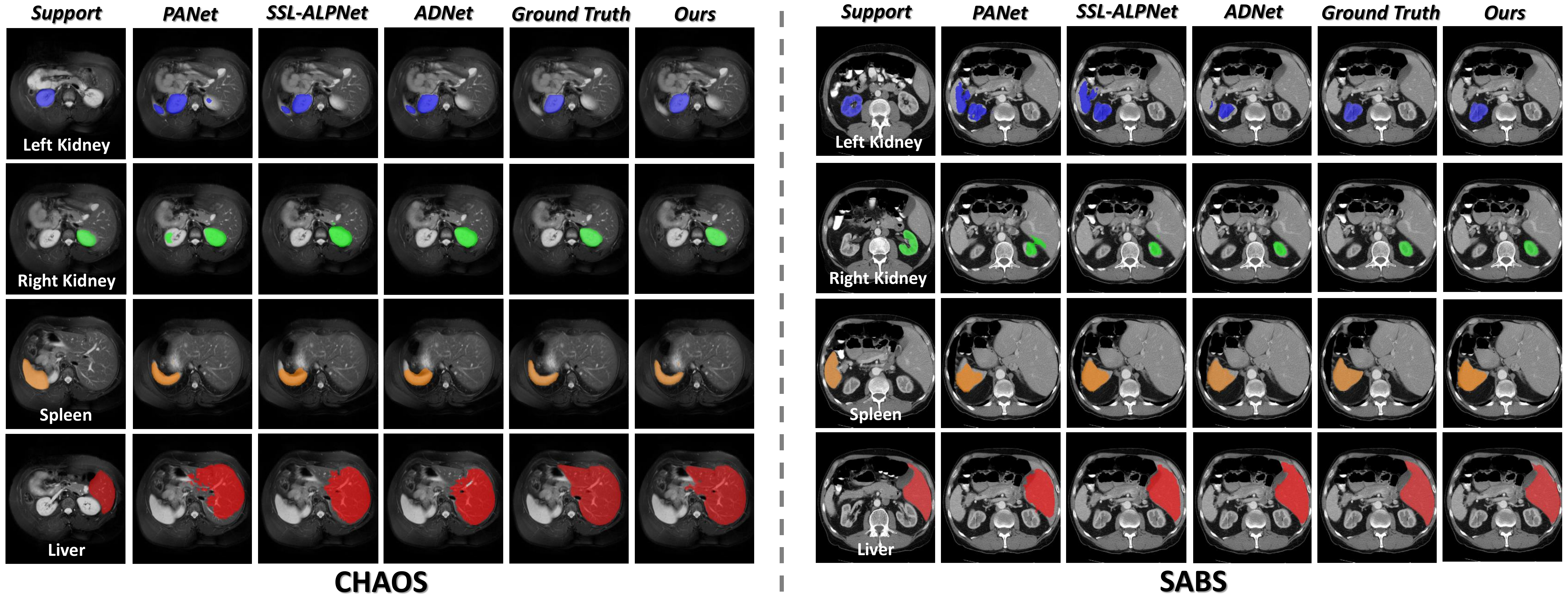}
\vspace{-2ex}
\caption{Qualitative results of our model on Abd-MRI and Abd-CT.}
\label{fig3}
\end{figure}

\subsection{Quantitative and Qualitative Results}
Table \ref{tab1} shows the performance comparison of the proposed method with state-of-the-art methods, including the vanilla PA-Net~\cite{wang2019panet}, SE-Net~\cite{roy2020squeeze}, ADNet~\cite{hansen2022anomaly}, CRAPNet~\cite{ding2023few}, SSL-ALPNet~\cite{ouyang2020self,ouyang2022self}, AAS-DCL~\cite{wu2022dual} and SR\&CL~\cite{wang2022few} under two experimental settings. From Tab. \ref{tab1}, it can be seen that the proposed method outperforms all listed methods in terms of the Mean values obtained under two different settings. Especially, the Mean value on Abd-CT dataset under Setting 1 reaches 77.83, which is 3.31 higher than the best result achieved by AAS-DCL. Consistent improvements are also indicated for Card-MRI dataset and can be found in the Appendix. In addition to the quantitative comparisons, qualitative results of our model and the other model on Abd-MRI and Abd-CT are shown in Fig. \ref{fig3} (See Appendix for CMR dataset). It is not difficult to see that our model shows considerable bound-preserving and generalization capabilities. 

\begin{table}[!t]
\begin{minipage}{0.6\textwidth}
		\centering
		\includegraphics[scale=0.18]{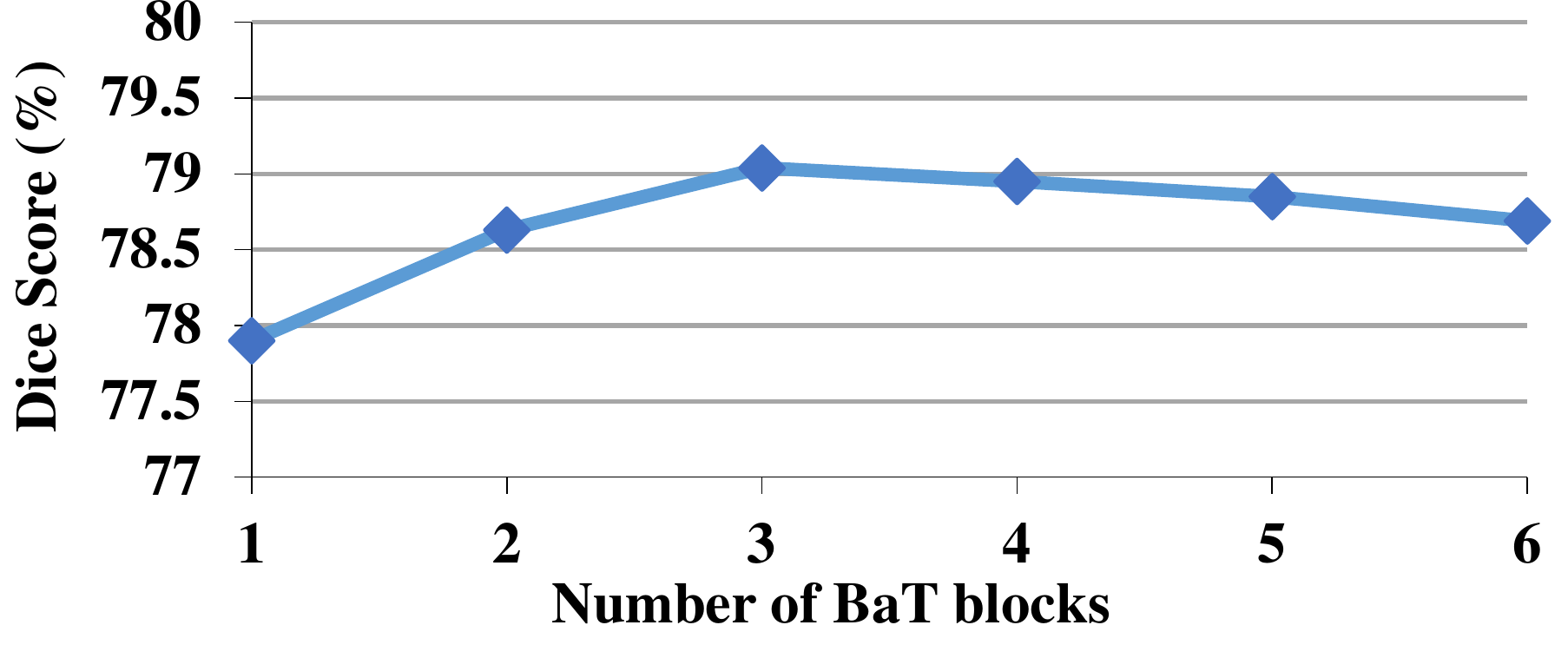}
		\makeatletter\def\@captype{figure}\makeatother\caption{Analysis of the number of BaT blocks.}
		\label{fig4}		
	\end{minipage} \hfill
	\begin{minipage}{0.4\textwidth}
		\centering
  \vspace{-3ex}
        \caption{Ablation study of the three loss functions.}
        \label{tab5}
        \begin{tabular}{ccc|c}
        \hline
        $\mathcal{L}_{ce}$& $\mathcal{L}_{B}$ & $\mathcal{L}_{dice}$& Dice score   \\ \hline
        \checkmark &  &  & 78.43 \\
        \checkmark & \checkmark &  & 78.81 \\
        \checkmark & \checkmark & \checkmark & \textbf{79.04} \\ \hline
        \end{tabular}
	\end{minipage}
 \vspace{-2ex}
\end{table}

\subsection{Ablation Studies}
The ablation studies were conducted on Abd-MRI dataset under Setting 2. As can be seen from Fig. \ref{fig4}, the use of three stacked BaT blocks is suggested to obtain the best Dice score. From Tab. \ref{tab5}, using a combination of boundary and dice loss gives a 0.61 increase in terms of the dice score compared to using only the cross-entropy loss. More ablation study results can be found in Appendix.

\section{Conclusion}
In this paper, we introduced a Region-enhanced Prototypical Transformer (RPT) to mitigate the impact of large intra-class variations present in medical image segmentation. The model is mainly beneficial from a subdivision-based strategy used for generating a set of regional support prototypes and a self-selection mechanism introduced to the Bias-alleviated Transformer (BaT) blocks. The proposed RPT can iteratively optimize the generated regional prototypes and output a more precise global prototype for predictions. The results of extensive experiments and ablation studies can demonstrate the advancement and effectiveness of the proposed method.

\bibliographystyle{splncs04}
\bibliography{references}

\begin{thebibliography}{10}
\providecommand{\url}[1]{\texttt{#1}}
\providecommand{\urlprefix}{URL }
\providecommand{\doi}[1]{https://doi.org/#1}

\bibitem{aurenhammer1991voronoi}
Aurenhammer, F.: Voronoi diagrams: a survey of a fundamental geometric data
  structure. ACM Computing Surveys (CSUR)  \textbf{23}(3),  345--405 (1991)

\bibitem{deng2009imagenet}
Deng, J., Dong, W., Socher, R., Li, L.J., Li, K., Fei-Fei, L.: Imagenet: A
  large-scale hierarchical image database. In: 2009 IEEE conference on computer
  vision and pattern recognition. pp. 248--255. Ieee (2009)

\bibitem{ding2023few}
Ding, H., Sun, C., Tang, H., Cai, D., Yan, Y.: Few-shot medical image
  segmentation with cycle-resemblance attention. In: Proceedings of the
  IEEE/CVF Winter Conference on Applications of Computer Vision. pp. 2488--2497
  (2023)

\bibitem{feng2021interactive}
Feng, R., Zheng, X., Gao, T., Chen, J., Wang, W., Chen, D.Z., Wu, J.:
  Interactive few-shot learning: Limited supervision, better medical image
  segmentation. IEEE Transactions on Medical Imaging  \textbf{40}(10),
  2575--2588 (2021)

\bibitem{hansen2022anomaly}
Hansen, S., Gautam, S., Jenssen, R., Kampffmeyer, M.: Anomaly
  detection-inspired few-shot medical image segmentation through
  self-supervision with supervoxels. Medical Image Analysis  \textbf{78},
  102385 (2022)

\bibitem{he2016deep}
He, K., Zhang, X., Ren, S., Sun, J.: Deep residual learning for image
  recognition. In: Proceedings of the IEEE conference on computer vision and
  pattern recognition. pp. 770--778 (2016)

\bibitem{kavur2021chaos}
Kavur, A.E., Gezer, N.S., Bar{\i}{\c{s}}, M., Aslan, S., Conze, P.H., Groza,
  V., Pham, D.D., Chatterjee, S., Ernst, P., {\"O}zkan, S., et~al.: Chaos
  challenge-combined (ct-mr) healthy abdominal organ segmentation. Medical
  Image Analysis  \textbf{69},  101950 (2021)

\bibitem{kervadec2019boundary}
Kervadec, H., Bouchtiba, J., Desrosiers, C., Granger, E., Dolz, J., Ayed, I.B.:
  Boundary loss for highly unbalanced segmentation. In: International
  conference on medical imaging with deep learning. pp. 285--296 (2019)

\bibitem{ABD-CT}
Landman, B., Xu, Z., Igelsias, J., Styner, M., Langerak, T., Klein, A.: Miccai
  multi-atlas labeling beyond the cranial vault--workshop and challenge. In:
  Proceedings of MICCAI Multi-Atlas Labeling Beyond Cranial Vault—Workshop
  Challenge. vol.~5, p.~12 (2015)

\bibitem{lin2014microsoft}
Lin, T.Y., Maire, M., Belongie, S., Hays, J., Perona, P., Ramanan, D.,
  Doll{\'a}r, P., Zitnick, C.L.: Microsoft coco: Common objects in context. In:
  European Conference on Computer Vision. pp. 740--755 (2014)

\bibitem{liu2020prototype}
Liu, J., Qin, Y.: Prototype refinement network for few-shot segmentation. arXiv
  preprint arXiv:2002.03579  (2020)

\bibitem{ma2021loss}
Ma, J., Chen, J., Ng, M., Huang, R., Li, Y., Li, C., Yang, X., Martel, A.L.:
  Loss odyssey in medical image segmentation. Medical Image Analysis
  \textbf{71},  102035 (2021)

\bibitem{ouyang2020self}
Ouyang, C., Biffi, C., Chen, C., Kart, T., Qiu, H., Rueckert, D.:
  Self-supervision with superpixels: Training few-shot medical image
  segmentation without annotation. In: European Conference on Computer Vision.
  pp. 762--780 (2020)

\bibitem{ouyang2022self}
Ouyang, C., Biffi, C., Chen, C., Kart, T., Qiu, H., Rueckert, D.:
  Self-supervised learning for few-shot medical image segmentation. IEEE
  Transactions on Medical Imaging  \textbf{41}(7),  1837--1848 (2022)

\bibitem{roy2020squeeze}
Roy, A.G., Siddiqui, S., P{\"o}lsterl, S., Navab, N., Wachinger, C.: `squeeze
  \& excite' guided few-shot segmentation of volumetric images. Medical image
  analysis  \textbf{59},  101587 (2020)

\bibitem{shen2022q}
Shen, Q., Li, Y., Jin, J., Liu, B.: Q-net: Query-informed few-shot medical
  image segmentation. arXiv preprint arXiv:2208.11451  (2022)

\bibitem{shen2021poissonseg}
Shen, X., Zhang, G., Lai, H., Luo, J., Lu, J., Luo, Y.: Poissonseg:
  semi-supervised few-shot medical image segmentation via poisson learning. In:
  IEEE international conference on Bioinformatics and biomedicine. pp.
  1513--1518 (2021)

\bibitem{snell2017prototypical}
Snell, J., Swersky, K., Zemel, R.: Prototypical networks for few-shot learning.
  Advances in neural information processing systems  \textbf{30} (2017)

\bibitem{sun2022few}
Sun, L., Li, C., Ding, X., Huang, Y., Chen, Z., Wang, G., Yu, Y., Paisley, J.:
  Few-shot medical image segmentation using a global correlation network with
  discriminative embedding. Computers in biology and medicine  \textbf{140},
  105067 (2022)

\bibitem{wang2019panet}
Wang, K., Liew, J.H., Zou, Y., Zhou, D., Feng, J.: Panet: Few-shot image
  semantic segmentation with prototype alignment. In: proceedings of the
  IEEE/CVF international conference on computer vision. pp. 9197--9206 (2019)

\bibitem{wang2022few}
Wang, R., Zhou, Q., Zheng, G.: Few-shot medical image segmentation regularized
  with self-reference and contrastive learning. In: International Conference on
  Medical Image Computing and Computer-Assisted Intervention. pp. 514--523
  (2022)

\bibitem{wu2022dual}
Wu, H., Xiao, F., Liang, C.: Dual contrastive learning with anatomical
  auxiliary supervision for few-shot medical image segmentation. In: European
  Conference on Computer Vision. pp. 417--434 (2022)

\bibitem{zhang2022feature}
Zhang, J.W., Sun, Y., Yang, Y., Chen, W.: Feature-proxy transformer for
  few-shot segmentation. In: Advance in Neural Information Processing Systems
  (2022)

\bibitem{zhuang2018multivariate}
Zhuang, X.: Multivariate mixture model for myocardial segmentation combining
  multi-source images. IEEE transactions on pattern analysis and machine
  intelligence  \textbf{41}(12),  2933--2946 (2018)

\end{thebibliography}

\end{document}